\pdfoutput=1
\documentclass[11pt]{article}
\usepackage[preprint]{acl}

\usepackage{times}
\usepackage{latexsym}
\usepackage[T1]{fontenc}
\usepackage[utf8]{inputenc}
\usepackage{microtype}
\usepackage{inconsolata}
\usepackage{graphicx}
\usepackage{fontawesome} 

\title{TextClass Benchmark: A Continuous Elo Rating of LLMs in Social Sciences\thanks{All the materials related to the TextClass Benchmark project are readily available on the {\faGithub} \href{https://github.com/bgonzalezbustamante/TextClass-Benchmark}{GitHub repository}, ensuring easy access for interested parties. In addition, the continuous Elo rating and Meta-Elo are displayed on the project’s web interface {\faHome} \href{https://textclass-benchmark.com/}{https://textclass-benchmark.com}, providing real-time updates and insights.}}

\author{Bastián González-Bustamante\thanks{Post-doctoral Researcher in Computational Social Science, Institute of Public Administration, Faculty of Governance and Global Affairs, Leiden University, Netherlands. {\faMapMarker} Wijnhaven, Turfmarkt 99, The Hague 2511 DP, Netherlands. Lecturer, School of Public Administration, Faculty of Administration and Economics, Universidad Diego Portales, Chile. {\faHome} \href{https://bgonzalezbustamante.com/}{https://bgonzalezbustamante.com}, ORCID iD \href{https://orcid.org/0000-0003-1510-6820}{https://orcid.org/0000-0003-1510-6820}.} \\
  Leiden University, Netherlands \\
  Universidad Diego Portales, Chile \\
  \normalsize{\href{mailto:b.a.gonzalez.bustamante@fgga.leidenuniv.nl}{b.a.gonzalez.bustamante@fgga.leidenuniv.nl}}}

\begin{document}
\maketitle
\begin{abstract}
The TextClass Benchmark project is an ongoing, continuous benchmarking process that aims to provide a comprehensive, fair, and dynamic evaluation of LLMs and transformers for text classification tasks. This evaluation spans various domains and languages in social sciences disciplines engaged in NLP and text-as-data approach. The leaderboards present performance metrics and relative ranking using a tailored Elo rating system. With each leaderboard cycle, novel models are added, fixed test sets can be replaced for unseen, equivalent data to test generalisation power, ratings are updated, and a Meta-Elo leaderboard combines and weights domain-specific leaderboards. This article presents the rationale and motivation behind the project, explains the Elo rating system in detail, and estimates Meta-Elo across different classification tasks in social science disciplines. We also present a snapshot of the first cycle of classification tasks on incivility data in Chinese, English, German and Russian. This ongoing benchmarking process includes not only additional languages such as Arabic, Hindi, and Spanish but also a classification of policy agenda topics, misinformation, among others.
\end{abstract}
\pagebreak

\section{Introduction}

The ability to work with and process large volumes of data is changing not only the landscape of the social sciences but also the humanities. Computational social sciences have gained ground in several disciplines, while the humanities have coined the digital humanities concept. In this context, rapid advances in machine learning and generative AI since the early 2020s are radically changing the research landscape, especially in the field of NLP and text-as-data. The accelerated pace in recent years has left slightly outdated machine learning techniques and text-as-data analysis focused on topic modelling, dictionaries and supervised or unsupervised approaches \citep[see also \citealp{GonzalezBustamante2023}]{Watanabe2022}. Even the BERT family, including fine-tuned or distilled BERT and roBERTa that have been used for several tasks in disciplines like political science \cite[see][]{Timoneda2024}, seems to pale in comparison to the rise of LLMs, in particular from GPT-4 and the Llama 3 architecture onwards.

Indeed, in several social science disciplines, LLMs have emerged not only as a new methodological tool but also as a sort of obsession. Some of the most well-known models are OpenAI’s GPTs, which include the novel o1-preview and o1-mini, released in September 2024 and were out of preview in early December 2024. These models have not only started to be used extensively in various tasks almost daily but they are also being used, via the OpenAI’s API, for various classification tasks and synthetic samples creation for research, thus replacing manual processes and conventional NLP approaches in several social science applications \citep{Argyle2023, Gilardi2023, GonzalezBustamante2024, Gruber2024, He2024, Linegar2023}.

However, this use is not without concerns. On the one hand, underlying biases in the training process of these models may influence the results they provide \citep{Geng2024, GonzalezBustamante2024}. There are concerns, on the other hand, related to the reliance on proprietary or for-profit models. These concerns relate to ethical considerations about transferring and using information without consent during training processes and the level of reproducibility these models offer. For this reason, open-source models have emerged as an alternative to collaborative research \citep{Spirling2023, Weber2023}.

Despite the concerns, deploying open-source models locally can be more complex than using GPTs through the OpenAI’s API. We used the term locally since several APIs of different providers, such as Mistral or Fireworks, allow the deployment of open-source models similar to OpenAI’s API. This option is beneficial for fine-tuning jobs or deploying models beyond RAM local infrastructure, such as Llama 3.1 405B parameters. Indeed, the API pay-per-use form offers resources beyond those generally available to average researchers in various social science fields, being simple and easy to implement without excessive computational requirements \citep{GonzalezBustamante2024, Linegar2023}.

In addition, this changing landscape and the variety of possibilities pose a challenge for generative AI in research: maintain the reproducibility of tasks performed using LLMs. Indeed, temperature experiments tend to show reproducibility issues \citep{Hao2024}, and it seems these models are more exposed to failed deterministic replication than annotation with crowdworkers \citep{Barrie2024b}. In this sense, in a field that appears to be rebuilding itself daily, clear standards are absent, however, some recommendations that are emerging highlight considering local deployments, prioritise open-source models, checking prompt strategies stability and running classification routines multiple times over time \citep{Barrie2024a, Barrie2024b}.

The TextClass Benchmark project is dedicated to testing the stability of a number of LLMs over time on different classification tasks. It aims to provide a comprehensive, fair, and dynamic evaluation of LLMs and transformers for text classification across various domains and languages in social sciences disciplines engaged in NLP and text-as-data approach. The project will incorporate prompt checks and offer insight into reproducibility issues and cross-model comparisons between closed and open-source LLMs. 

The following section provides a detailed description of the Elo rating system that we use in each cycle. We then describe the classification task and data used in the first cycle in Chinese, English, German, and Russian toxicity detection before presenting the results of this first snapshot. Finally, we briefly discuss some good practices for maintaining the project and future avenues.

\section{Elo Rating System}

\subsection{Elo Rating Overview}

The Elo system ---widely used in chess and a number of competitions--- allows us to benchmark dynamically different models and track relative performance over time. We used a baseline of 1,500 points for each model incorporated. Then, we ran pairwise comparisons between models in round-robin matches in each cycle. This implies that models are randomly paired, and each “plays” against another, considering their prediction performance on ground-truth evaluation using a fixed test data set. 

We estimate expected scores for each model pair A and B, with ratings $R_{A}$ and $R_{B}$ using the following standard formula borrowed from the classic proposal of the Hungarian-American physics, Arpad E. Elo:

\begin{equation}
    E_{A} = \frac{1}{1 + 10^{(R_{B} - R_{A}) / 400}} 
\end{equation}

\begin{equation}
    E_{B} = 1 - E_{A}
\end{equation}

Then, the F1-Score determines the winner because it is our primary absolute performance metric as a harmonic measure that combines precision and recall. However, the result is determined by margin-based comparison. Thus, if the difference in F1-Score between the two models is greater than 0.05, the model with the higher metric is the winner. This implies that the outcome is considered a draw in cases where the difference is within 0.05.

After all the matches, the rating is updated using the expected scores explained above and the actual outcome ($S_{i} = 1$ for win; $S_{i} = 0.5$ for draw; $S_{i} = 0$ for loss). Therefore, new ratings are calculated as follows:

\begin{equation}
    \hat{R}_{A} = R_{A} + K \times (S_{A} - E_{A})
\end{equation}

\begin{equation}
    \hat{R}_{B} = R_{B} + K \times (S_{B} - E_{B})
\end{equation}

Our $K$-Factor value is 40, which is relatively high since we want to generate quick adjustments in iterations and new cycles to reflect the performance of state-of-the-art models in the current research landscape with a high pace of generative AI progress.

\subsection{Meta-Elo}

In addition, we combined domain-specific Elo leaderboards controlling for classification task complexity, language data scarcity, absolute performance and cycle count. Therefore, we calculate the Meta-Elo indicator in the following manner:

\begin{equation}
    M_{i} = \sum_{j = 1}^{n} w_j \times R_{i[j]}
\end{equation}

We weigh each leaderboard as follows:

\begin{equation}
    w_{j} = w_{task} \times w_{language} \times w_{F1} \times w_{cycle}
\end{equation}

First, we measure task complexity as the logarithmic of the number of categories in the classification task plus one. Then, we assign higher weights to languages with lower digitalisation and data availability. We consider English a baseline and assign values such as 1.3 to Chinese, 1.1 to German and 1.4 to Russian.\footnote{These weights are trying to reflect not only language resource scarcity for NLP but also linguistic complexity and morphological challenges. The ongoing cycles also test Arabic, Hindi, and Spanish, with weights of 1.5, 1.7, and 1.2, respectively.}

We also consider absolute performance by incorporating a normalised F1-Score as weight by dividing it by the maximum F1-Score across models and leaderboards. Finally, we incorporate a weight that increases with the number of cycles as $1 + log(cycle + 1)$. The rationale for incorporating the number of cycles is to reward models that have been consistently benchmarked over several iterations instead of penalising fewer active models in a way to account for potential obsolescence. In this way, we also prevent a penalty on less-tested models because of deployment challenges in terms of costs, infrastructure or computing time.\footnote{Our current infrastructure allows us to deploy locally within the range between 70 and 100B parameters and through APIs OpenAI’s GPTs and Llama 3.2 405B parameters.}

It is important to bear in mind that both Elo scores are relative measures that focus on the comparative strengths of models. For this reason, it is relevant to consider absolute performance measures to have a clearer picture, such as the F1-Score in the case of Elo-Score. We adjusted a weighted F1-Score across leaderboards for Meta-Elo, emulating the abovementioned process.

\section{Task Description}

This paper presents the models tested in the toxicity classification first cycle in Chinese, English, German, and Russian. We have used a balanced sample of 5,000 observations per country ($N$ = 20,000) split in a 70/15/15 proportion for training, validation, and testing in case of potential future fine-tuning jobs during the subsequent cycles. The data correspond to several sources used in the framework of the Multilingual Text Detoxification \citep[TextDetox, 2024, see][]{Dementieva2024}. This shared task wanted to promote a proactive approach to online toxicity by presenting a neutral version of the messages that maintains the content’s meaning. For the text detoxification challenge, a number of sources were used comprising toxic and nontoxic messages, for example, Jigsaw and Unitary AI toxicity Wikipedia data \citep[see][]{Hanu2020} for English, DeTox-Dataset \citep[see][]{Demus2022} and GemEval \citep[see][]{Risch2021} with Twitter and Facebook comments for German, among other sources.

Our task involved zero-shot binary classification using Google’s and Jigsaw’s core definition of incivility, similar to the prompt strategy by \citet{GonzalezBustamante2024}: {\itshape “Classify the category of the comment as either TOXIC or NONTOXIC. TOXIC: Rude, disrespectful, or unreasonable comments that are likely to make someone leave the discussion or stop sharing their perspective. NONTOXIC: Civil or nice comments that are unlikely to discourage conversation”}. The temperature was set at zero, and the performance metrics were averaged for binary classification. In addition, other relevant LLMs parameters, such as repeat penalty, nucleus and top-$k$ sampling, and minimum probability for token selection, were adjusted carefully at the standard values of Ollama.\footnote{We only altered the random number for text generation, however, we used the same number for all models.}

\begin{figure*}[t]
  \includegraphics[width=0.99\linewidth]{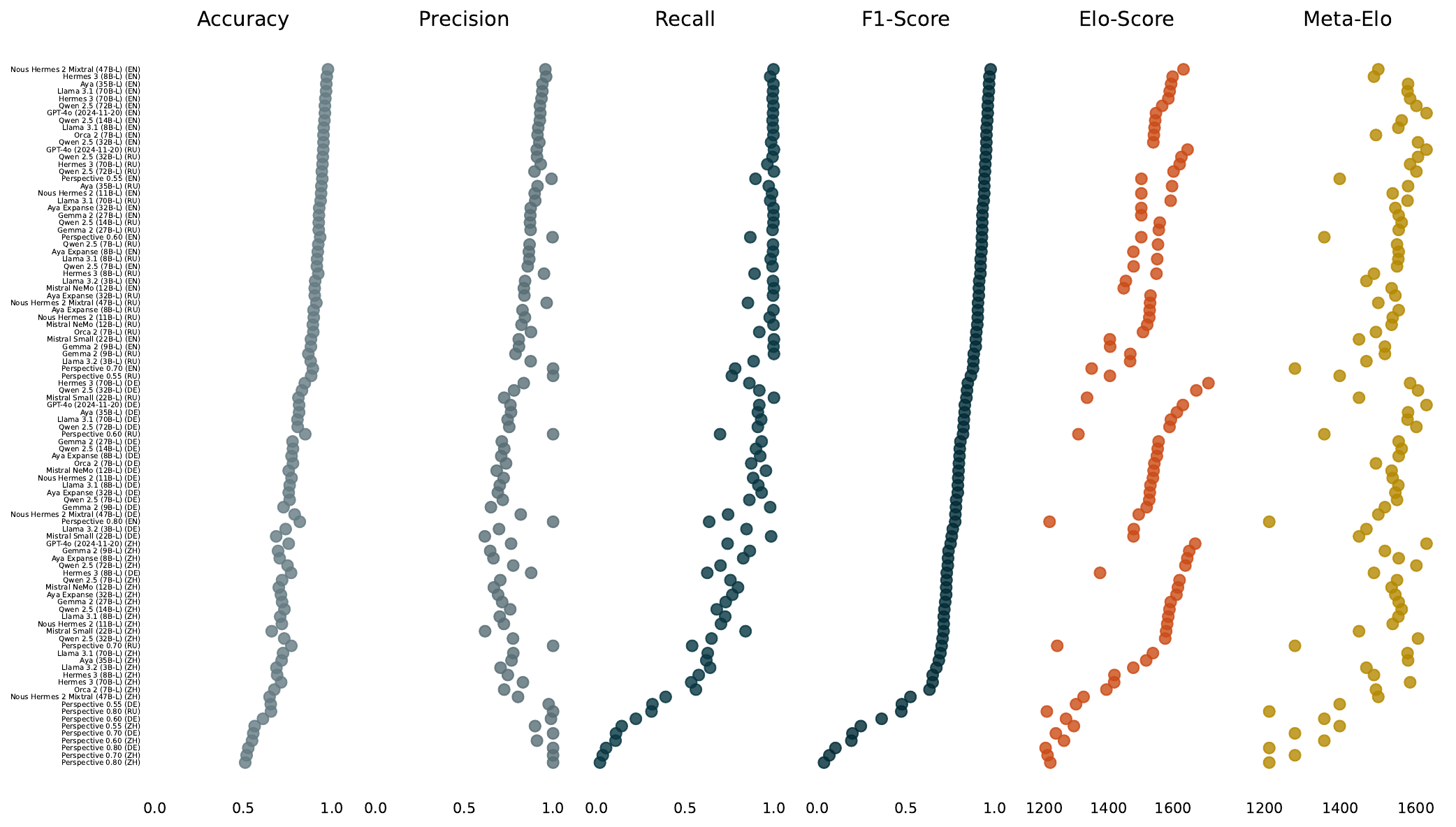}
  \caption {Goodness-of-Prediction Metrics, Elo-Score, and Meta-Elo \\ {\footnotesize {\itshape Note}. Accuracy represents the proportion of correct predictions among all the predictions made. Precision denotes the ratio of true positive predictions and reflects how much the model avoids Type I errors (false positives). Recall indicates the proportion of actual positive cases correctly predicted; it reflects how well the model avoids Type II errors (false negatives). The F1-Score combines precision and recall into a metric by calculating the harmonic mean. After the billions of parameters in parenthesis, the uppercase L implies that the model was deployed locally.}}
  \label{fig:performance}
\end{figure*}

This snapshot benchmarked one of the flagship models of OpenAI: GPT-4o (2024-11-20). We also tested the well-known Perspective API, a distilled BERT developed by Jigsaw and Google that was once cutting-edge but is now an off-the-shelf option for toxicity classification. Then, we focused on testing some relevant and ---for the moment--- state-of-the-art open-source LLMs deployed locally on a high-performance workstation with considerable GPU capacity: Aya Expanse 8B and 32B, Gemma 2 9B and 27B, Hermes 3 8B and 70B, Llama 3.1 8B and 70B, Llama 3.2 3B, Mistral NeMo 12B, Mistral Small 22B, almost all Qwen 2.5 (7B, 14B, 32B and 72B) and Solar Pro 22B. We also tested some slightly outdated open-source LLMs whose performance should be reasonable: Mistral OpenOrca 7B, Nous Hermes 2 11B, Nous Hermes 2 Mixtral 47B, Orca 2 7B.\footnote{The models that tend to self-promote their multilingual capabilities are Aya, Aya Expanse, GPTs, Llama, Perspective API ---only for toxicity detection--- and Qwen 2.5.}

\section{First Snapshot}

This very first snapshot presents 24 models tested a total of 96 times. We have weighted the classic performance metrics binary and estimated Elo-Score and Meta-Elo across leaderboards for toxicity classification in Chinese, English, German, and Russian. Figure \ref{fig:performance} presents all the metrics per model and language listed by the F1-Score in descending order. In this vein, it is relevant to note that both Elo-Score and Meta-Elo highlight comparative strengths, however, the classic goodness-of-prediction indicators show the absolute performance, especially F1-Score.

A visual inspection allows us to identify a hierarchy by language. Models tend to perform better in English (average F1-Score = 0.952) and Russian (average F1-Score = 0.910). Then, models in German (average F1-Score = 0.814) tend to considerably outperform the Chinese classification (average F1-Score = 0.346). One interesting case is Nous Hermes 2 Mixtral 47B parameters, trained on Mixtral over GPT-4 synthetic data, that outperforms all models when classifying English data (F1-Score = 0.977). However, it shows a poor performance in Chinese (F1-Score = 0.524). In German, Hermes 3 70B parameters outperforms all other models (F1-Score = 0.848), while GPT-4o performs best in both Russian (F1-Score = 0.952) and Chinese (F1-Score = 0.751). Another interesting finding is that all LLMs outperform more classical transformer approaches such as Perspective API.

Elo-Score allows us to visualise better the differences in languages since it tends to order the performance within the domain-specific leaderboards. Meta-Elo, on the other hand, tends to smooth the Elo breaks between leaderboards.

\begin{figure}[t]
  \includegraphics[width=0.99\linewidth]{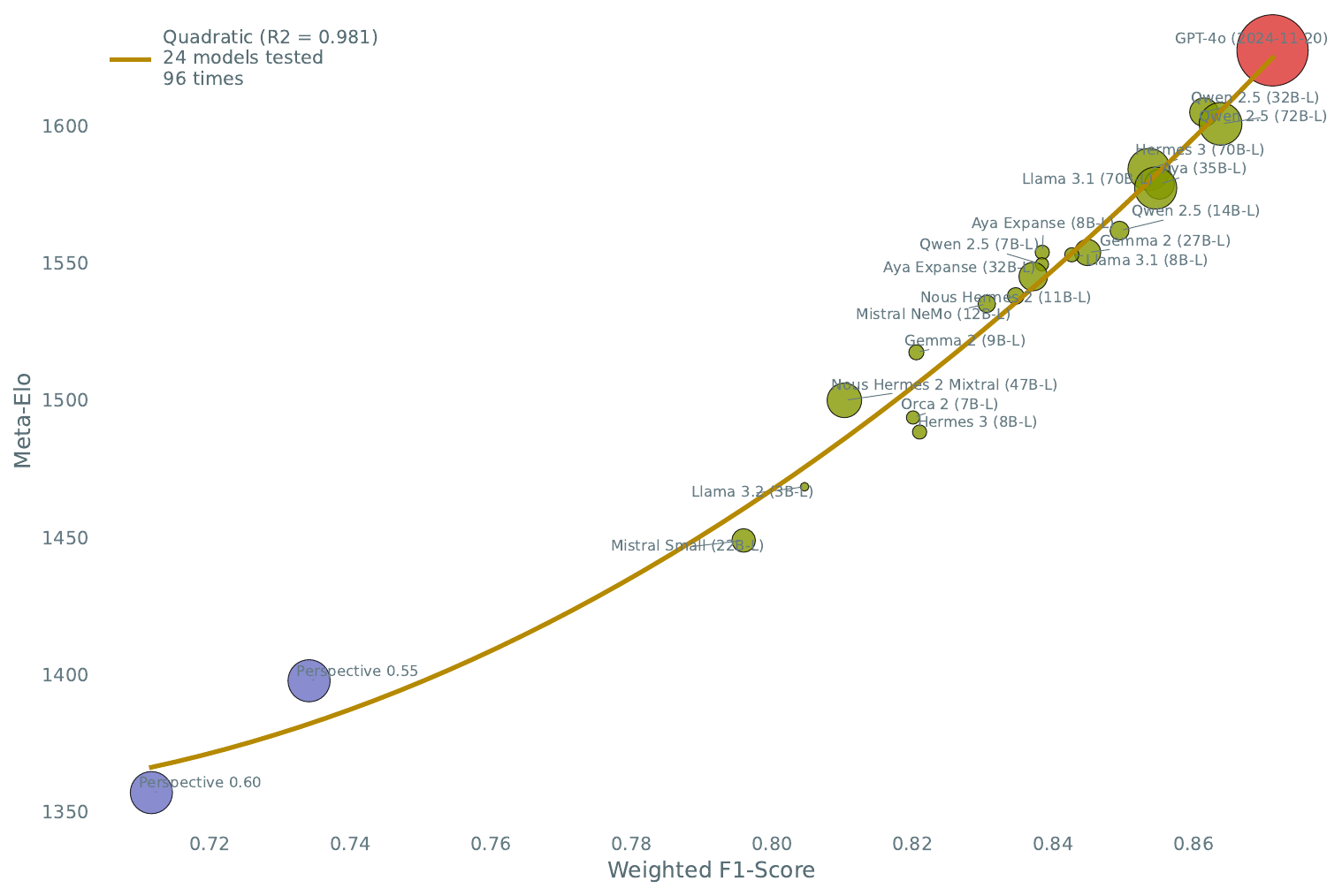}
  \caption {Weighted F1-Score and Meta-Elo \\ {\footnotesize {\itshape Note}. Green: open-source; red: closed-private; violet: BERT family. Models with a weighted F1-Score lower than 0.7 were excluded. Bubble size represents the ---presumed for GPTs--- number of parameters.}}
  \label{fig:meta_elo}
\end{figure}

Figure \ref{fig:meta_elo} presents the relationship between weighted F1-Score and Meta-Elo. It shows a quadratic relationship with a high goodness-of-fit ($R^{2}$ = 0.981). Across leaderboards, the best model in this first snapshot is GPT-4o (2024-11-20). In addition, the plot shows a strong relationship between absolute and relative performance measured by F1-Score and Meta-Elo, respectively. However, in the mid-term, we should be able to identify some deviances when adding new cycles, which should help not only to get a better understanding of the differences between Elo measures and goodness-of-prediction metrics but also to test the stability of LLMs classification over time.

\section{Discussion: Good Practices and New Avenues}

Even though LLMs are taking by storm the methodological landscape in almost every discipline, they are not a panacea. It is pretty evident in this first snapshot that they perform differently depending on the language and the task. That is, without considering the number of specifications that could affect the outcome and compromise reproducibility in annotation tasks, such as the well-known temperature parameter, the nucleus sampling for diversity or the token repetition penalty. To tackle this fragility, snapshots like this and the classic way to produce knowledge in social sciences and other fields, which involves slow peer-reviewing that delays results release to the point that the models and findings are obsolete, are insufficient. In this sense, a continuous test of LLMs should offer a better understanding of the performance of these models for specific tasks in social science research.

In this continuous benchmarking, we will maintain specific practices for handling deprecated models and data splits policy. First, we shall incorporate novel models in the mid-term and may deprecate some slightly outdated ones to control for obsolescence.\footnote{In the ongoing cycles, we already incorporated state-of-the-art models such as o1-preview, o1-mini, and Llama 3.1 405B, among others. In addition, we will soon incorporate o1 out of the preview and the novel Llama 3.3.} We will rely on a Keep the Last Known Elo-Score policy that maintains the rating’s integrity without match simulations. We will retain the inactive models’ Elo-Score from the last cycle they participated in and mark them inactive in the reports. Should the models re-enter in future cycles, their last Elo-Score will serve as a starting point, ensuring the continuity of our benchmark. The data splits policy, on the other hand, wants to provide a reliable and consistent model evaluation. Each model is evaluated on a fixed test set for each classification task in each cycle. This is particularly relevant for the potential incorporation of fine-tuned LLMs or BERT family models and allows for fair comparison using novel, unseen data to control overfitting, inflated metrics and lack of generalisation power. This policy is relevant because it also prevents data leakage and train-test contamination.

We also recognise that the research landscape in NLP and generative AI is continually evolving, therefore, we expect to update the wights associated with language data scarcity in order to reflect the latest trends and data availability. Similarly, we shall evaluate the number of cycles weight to control for obsolescence. Models with lower performance might receive disproportionate bonification, leading to overestimation. In this sense, we need to explore some log-sigmoid scaling methods but control specific interactions in the weights that could introduce subtle variability and generate a new weight that exceeds the current one.

If necessary, we shall document dual-test approaches in which we may use equivalent {\itshape pseudo}-test sets to estimate metrics to ensure that fine-tuned models do not recall learned patterns from training data. In addition, we will apply stratified sampling for imbalanced data to maintain the same proportion of labels across train, validation, and tests set when necessary. Subsequently, we will use proper averaging to estimate the absolute performance metrics.

Finally, with each leaderboard cycle, novel models shall be added, fixed test sets could be replaced for unseen, equivalent data to test generalisation power, and ratings will be updated. Although there are no fixed updates, we will update each leaderboard continuously by incorporating state-of-the-art and fine-tuned models and new data sources relevant to social science disciplines in order to offer insights into the stability of LLMs for a variety of relevant classification and annotation tasks.

%% \section*{Acknowledgements}

%% {\itshape Omitted in the review version.}

\bibliography{TextClass-Benchmark}

\appendix

\setcounter{table}{0}
\renewcommand{\thetable}{A\arabic{table}}

%% \section{Appendix}
%% \label{sec:appendix}

\end{document}